\documentclass[letterpaper]{article}

\usepackage[margin=0.5in]{geometry} 
\usepackage{array}                
\usepackage{graphicx}             
\usepackage{booktabs}             
\usepackage{arydshln}             
\usepackage{adjustbox}            
\usepackage{rotating}             
\usepackage{makecell}
\usepackage{multirow}

\usepackage{times,amsmath,epsfig}
\usepackage{fancyhdr}
\usepackage{amsmath}
\usepackage{amsfonts}
\usepackage{amssymb}
\usepackage{array}
\usepackage{graphicx}
\usepackage{url}
\usepackage{subfigure}
\usepackage{bm}
\usepackage{breqn}
\usepackage{xcolor}
\usepackage{soul}
\usepackage{amssymb}
\usepackage[breaklinks=true]{hyperref}
\usepackage{breakcites}
\usepackage{microtype}

\begin{document}

\vspace{1 cm}

\title{Robust Computer-Vision based Construction Site Detection for Assistive-Technology Applications}

%

\author{ Junchi Feng, Giles Hamilton-Fletcher, Nikhil Ballem, Michael Batavia, Yifei Wang, Jiuling Zhong, \\ Maurizio Porfiri, John-Ross Rizzo
}

\author{Junchi Feng$^{a,b}$,
         Giles~Hamilton-Fletcher$^{c,d}$, 
        Nikhil Ballem$^d$, 
        Michael Batavia$^d$, 
        Yifei Wang$^b$,
        Jiuling Zhong$^b$,
        \\
        Maurizio~Porfiri$^{a,b,e}$,
        John-Ross~Rizzo$^{a,c,d}$
}


\date{}
\maketitle

\textbf{Affiliations:}
\begin{itemize}
    \renewcommand{\labelitemi}{}
    \item $^a$ Department of Biomedical Engineering, Tandon School of Engineering, New York University, Brooklyn, NY 11201, USA
    \item $^b$ Center for Urban Science and Progress, Tandon School of Engineering, New York University, Brooklyn, NY 11201, USA
    \item $^c$ Department of Ophthalmology, NYU Grossman School of Medicine, New York, NY 10016, USA
    \item $^d$ Department of Rehabilitation Medicine, NYU Grossman School of Medicine, New York, NY 10016, USA
    \item $^e$ Department of Mechanical and Aerospace Engineering, Tandon School of Engineering, New York University, Brooklyn, NY 11201, USA
\end{itemize}

\clearpage

\textbf{\Huge Robust Computer-Vision based Construction Site Detection for Assistive-Technology Applications}

%


\maketitle

\begin{abstract}
 
Navigating urban environments poses significant challenges for people with disabilities, particularly those with blindness and low vision. Environments with dynamic and unpredictable elements like construction sites are especially challenging. Construction sites introduce hazards like uneven surfaces, obstructive barriers, hazardous materials, and excessive noise, and they can alter routing, complicating safe mobility. Existing assistive technologies are limited, as navigation apps do not account for construction sites during trip planning, and detection tools that attempt hazard recognition, struggle to address the extreme variability of construction paraphernalia.

This study introduces a novel computer vision–based system that integrates open-vocabulary object detection, a YOLO-based scaffolding-pole detection model, and an optical character recognition (OCR) module to comprehensively identify and interpret construction site elements for assistive navigation. In static testing across seven construction sites, the system achieved an overall accuracy of 88.56\%, reliably detecting objects from 2m to 10m within a 0°–75° angular offset. At closer distances (2–4m), the detection rate was 100\% at all tested angles. At 10m, six of the seven sites remained detectable between 0° and 15°. In a dynamic test involving a 0.5-mile walking route, the system demonstrated 91.0\% accuracy in distinguishing construction from non-construction frames. These results suggest that the proposed system can significantly enhance navigation safety for individuals with blindness or low vision.

\end{abstract}


\section{Introduction}

Metropolitan areas are home to approximately two-thirds of adults with disabilities in the United States, highlighting their role as critical locations that require accessibility and tailored support \cite{zhao2019prevalence}. These urban centers serve as access hubs, offering extensive public transportation systems, inclusive sidewalks, and essential services \cite{zhao2019prevalence}. Despite these advantages, significant challenges remain for the independent mobility of people with disabilities, and particularly for persons with blindness and low vision (pBLV). One significant challenge is the first-and-last-mile problem — the difficulty of navigating to and from transit stops \cite{mohiuddin2021planning}, a short journey that may be riddled with oddly placed street furniture, inconvenient crowds, parked cars on pavements, inaccessible pedestrian crossings, and hazardous construction sites \cite{rnib2015who}. Construction sites and related paraphernalia introduce dynamic and unpredictable barriers like uneven surfaces, hazardous materials, excessive noise, and altered sidewalk layouts, which complicate safe navigation by undermining spatial memory and route familiarity \cite{matsuda2021gazing}.

Unfortunately, construction is also a ubiquitous feature of urban environments \cite{denis2020more}, as cities constantly expand, maintain, and upgrade their infrastructure. For example, New York City (NYC) currently has over 1,100 active construction sites \cite{NYC_Housing_Development_Active_Projects_2024} and 9,000 sidewalk sheds stretching more than 400 miles \cite{NYC_Mayor_2024}. Construction sites present various hazards that complicate navigation for those with disabilities. One of the primary challenges is the temporary alterations in terrain, such as uneven surfaces, gravel, trenches, or cable protectors, which are particularly difficult to detect with traditional mobility aids like white canes and challenging to navigate using wheelchairs \cite{husin2020inwalker}. These mobility aids are not designed to detect subtle ground changes, increasing the risk of accidents. In addition to hazardous terrain, construction sites frequently contain obstructive barriers, misplaced equipment, and scattered debris that can disrupt intended pathways. Unpredictable barriers make it difficult for pBLV to assess the area's layout, increasing the risk of disorientation and accidents. Construction site noise presents additional challenges for pBLV, whereby loud machinery and ongoing construction activities generate noise that can mask essential auditory cues to safe navigation, such as traffic sounds and pedestrian alerts. This noise can also drown out the audio feedback of assistive technologies, making it difficult for users to hear important navigational cues clearly \cite{brewster1998using,strumillo2018different}. These obstacles underscore the urgent need for assistive technologies that can reliably detect construction sites and help users either navigate around them or pass through safely.

Existing assistive technology-based navigation tools often fall short in adequately addressing the temporary but significant disruptions faced by travelers with disabilities. Most trip-planning applications, such as Google Maps \cite{gibson2006google}, prioritize providing the most efficient routes but do not necessarily account for safety and accessibility for pBLV. While Google Maps effectively provides routing and directions, its algorithm mainly focuses on traffic conditions, road closures, and other large-scale disruptions \cite{lanning2014dijkstra}. It may not always have the latest information on construction sites unless these are major projects that impact general traffic flow significantly. Smaller construction activities, especially those affecting pedestrian pathways rather than roadways, might not be reported promptly or accurately, making it difficult for pBLV to find the safest and most accessible routes.

Similarly, current computer vision-based methods for detecting construction sites and related objects, which typically utilize CNN-based frameworks \cite{fang2018automated, wang2021fast, xiao2021development}, depend heavily on annotated datasets. While these frameworks are generally adept at recognizing certain object types, their reliance on annotated datasets poses a significant hurdle. The effectiveness of these models is inherently tied to the diversity of object classes in the training data. As a result, they often struggle with unannotated or novel objects. Moreover, computer vision models for construction sites are usually developed for safety monitoring from video surveillance footage \cite{alateeq2023construction,khan2023construction}. By contrast, research focusing on detecting construction sites from a pedestrian's perspective is limited. These limitations highlight a critical need for innovation to support pBLV effectively. The generation of extensive annotated datasets specifically for construction environments is not only a labor-intensive process but also entails considerable costs  \cite{paneru2021computer}. Since these datasets are crucial for training detection models, their limitations can hinder the performance and reliability of navigation tools. Moreover, construction-related objects are regulated under city-specific standards, often varying in color or shape. Such nuances in object presentation can complicate detection efforts, particularly for computer vision methods trained on limited or non-diverse datasets.

To address these gaps, this study proposes a novel computer vision-based open vocabulary construction site detection model. Unlike traditional object detection systems that rely on predefined categories from annotated datasets, the proposed model leverages open vocabulary detection to identify and localize a wide range of construction objects based on textual descriptions. This capability is particularly valuable in the field of construction site detection, where the diversity of objects—such as tools, barriers, machinery, and signage—makes it impractical to rely solely on fixed object classes. By enabling the detection of previously unseen or unannotated objects, open vocabulary detection overcomes the limitations of dataset dependency, allowing for greater flexibility and adaptability to real-world scenarios. A YOLO model \cite{jiang2022review} is specifically trained to detect certain classes, serving as a supplement to the open vocabulary model to enhance accuracy.  Additionally, the integration of optical character recognition (OCR) enhances the system’s utility by enabling the detection of construction signs, providing additional context and warnings to users. This approach not only addresses the challenges posed by dynamic construction environments but also sets a precedent for leveraging open vocabulary techniques in other fields requiring adaptable object detection.

We hypothesize that integrating open vocabulary detection and OCR into a model for construction site detection can significantly improve the accuracy, range, and field-of-view of detection, as required for safe navigation in dynamic urban settings. To validate this hypothesis, we conducted a series of experiments aimed at evaluating our model’s accuracy, angular offset, and detection range.

\section{Methods}

\subsection{The Detection of Construction Sites}

A single construction site-related object is insufficient to confirm the presence of a construction site. We developed a comprehensive decision-making framework to afford the reliable inference of construction zones. This framework consists of three pipelines, each analyzing the image from a different perspective: sidewalk sheds, construction site-related objects, and construction signs.

\textbf{1. Sidewalk shed detection}: A commonly encountered type of construction site is the sidewalk shed. A sidewalk shed is a temporary structure installed over a sidewalk to protect pedestrians from falling debris during construction \cite{nyc_sidewalk_sheds}. Its main components include decking, vertical supports, and bracing \cite{nyc_project_categories_sidewalk_shed}. The deck, typically made of wood or steel, is the overhead platform providing protection. Vertical supports are upright poles holding up the decking, while bracing consists of diagonal or horizontal elements that stabilize the structure and prevent collapse. Since the deck has few distinctive visual features and is typically flat, object detection models struggle to recognize it. Our model identifies the presence of sidewalk sheds by detecting vertical supports and horizontal/diagonal bracing, typically steel poles in NYC \cite{nyc_project_categories_sidewalk_shed}. Theoretically, a sidewalk shed requires a minimum of four steel poles to form a stable rectangular frame (two on each side) and two diagonal braces to prevent swaying, resulting in at least six steel poles. To allow for some tolerance for false negatives, detecting at least five steel poles—either vertical supports or bracing—confirms the presence of a construction site.

\textbf{2. Construction-related objects detection}: The presence of construction-related objects, such as traffic barricades, barrels, cones, green wooden walls, and similar items, is often strong evidence of an active construction site. These objects secure the area, guide pedestrians, and facilitate construction activities. In most cases, construction sites contain multiple construction-related objects to meet essential safety and operational requirements.  Therefore, detecting at least three construction-related objects serves as a criterion for confirming a construction site. This threshold reflects the typical presence of multiple objects at construction sites while allowing some tolerance for false positives triggered by some random objects on the street.

\textbf{3. Construction sign detection}: Signage serves as strong evidence of a construction site. Construction signage refers to an informational sign used to warn, guide, or manage traffic in or around a construction zone. Such signs are difficult for pBLV to localize and interpret. A sign explicitly displaying text such as ``Construction Zone" is sufficient to confirm the presence of a construction site. Therefore, the detection of a single sign typically used in construction sites is enough to conclude the existence of a construction site.

\subsection{Sidewalk Shed Detection}

As previously mentioned, our model identifies the presence of sidewalk sheds by detecting vertical supports and horizontal or diagonal bracing, typically made of steel poles. However, detecting sidewalk shed poles using an open vocabulary approach is particularly challenging due to their visual similarity to other vertical structures commonly found in urban environments, such as streetlight poles, glass window frames, and traffic sign poles.

To address this challenge, we trained a YOLOv8 model specifically designed to detect sidewalk shed poles. Volunteers collected images of sidewalk sheds, emphasizing scaffolding poles as the primary structural elements. This effort resulted in a comprehensive dataset of 447 images captured in diverse urban settings and under varying lighting conditions to ensure broad representation for training. The dataset was divided into training and validation subsets in an 80/20 split. Two distinct classes were annotated: scaffolding poles, which are vertical poles essential for supporting sidewalk sheds, and horizontal scaffolding, which includes poles oriented horizontally or nearly horizontally to connect vertical poles.

In total, 2,297 annotations were made for horizontal scaffolding and 2,593 annotations for scaffolding poles, providing a solid foundation for model training. The YOLOv8 model was trained for 100 epochs on this dataset and achieved a mean Average Precision (mAP) of 0.72 at 50\% IoU (mAP@50) and 0.506 across IoU thresholds (mAP@50–95) on the sidewalk construction validation dataset. For inference, the confidence threshold was set to 0.25 and the IoU threshold to 0.7. Class-specific performance metrics revealed precision and recall values of 0.683 and 0.604 for horizontal scaffolding, and 0.789 and 0.717 for scaffolding poles. These results highlight the model’s robust detection capabilities, even in complex scaffolding setups.

\subsection{Construction-related Objects Detection}

We propose to use an open vocabulary object detection model to handle the variability of objects found at construction sites. Open vocabulary object detection enables the recognition of a broad array of objects without the necessity for extensive training on specific categories \cite{cheng2024yolo}. The model used for this project is YOLO-World. YOLO-World is an advanced object detection framework based on the YOLO series, designed for real-time open vocabulary object detection \cite{cheng2024yolo}. It integrates vision-language modeling to overcome the traditional YOLO limitation of detecting only predefined categories. The reason for selecting YOLO-World is due to the balance between speed and accuracy. It reaches high efficiency with 52.0 FPS on NVIDIA V100 GPUs and strong zero-shot detection performance with 35.4 average precision on the LVIS dataset \cite{cheng2024yolo}.

To tailor YOLO-World for construction detection, we curated a specialized vocabulary focused on common construction objects. Notably, we included multiple descriptors for the same object to account for variations in appearance due to lighting conditions and the model's sensitivity to color nuances. For instance, terms like ``green wall", ``dark green wall", and ``green construction wall" were used to describe construction site wall dividers that may appear differently under varying illumination. In an open vocabulary setting, any single textual description might not perfectly align with the learned visual features \cite{radford2021learning}. Supplying multiple descriptions increases the likelihood that at least one textual embedding will closely match the object’s visual representation.

Objects representing construction sites include traffic cones, traffic barricades, traffic barrels, and construction wall dividers. To detect these objects, the terminology used for the construction site objects were: ``traffic cone", ``orange and white striped traffic barrier", ``construction barricade", ``white traffic barrier", ``red traffic barrier", ``traffic barrier", ``red traffic barrier", ``orange traffic barrier", ``red traffic barricade", ``white traffic barricade", ``red and white barricade", ``green construction wall", ``construction wall", ``green wall", ``dark green wall".

A prominent issue in open-vocabulary object detection is overgeneralization, which occurs when a model extends its decision boundary too broadly, inadvertently including areas of the feature space that do not correspond to any recognized class \cite{scheirer2012toward}.  This issue can lead to incorrect classifications where the model mistakenly labels objects with similar features as belonging to the same class. For example, in the case of detecting a ``white traffic barricade", if a white car is the only white object present, the model may erroneously classify the car as the barricade due to the overly generalized decision boundary.

To mitigate the problem of overgeneralization, we introduced null classes. A null class is defined as a category that the model is not specifically interested in detecting, but which can help improve the performance for classes of interest \cite{yoloworld_prompting_tips}. Incorporating these additional classes refines the model’s decision boundaries, improving its ability to distinguish visually similar objects. Our null classes were identified through testing with various images and common misclassifications were tracked. For example, objects such as ``fire hydrant" which was frequently misclassified as ``traffic cone" and ``green construction wall," often mistaken for ``grassland" were included as null classes to enhance model differentiation capabilities.

The following null classes were integrated into the model based on empirical testing to address common misclassifications: ``car", ``white car", ``truck", ``bench", ``fire hydrant", ``computer monitor", ``tree", ``tree canopy", ``building", ``grass", and ``grassland".

Moreover, customized confidence scores tailored to specific terms were used to improve detection accuracy in object detection models. A key characteristic of YOLO-World is its higher confidence levels for terms corresponding to classes in the training dataset \cite{yolo_world_prompting_tips}. To ensure accurate predictions for each term, adjusting confidence thresholds at the term level is important.

To manage variability in confidence scores, we established four distinct categories through internal testing. Each category addresses different classes of objects, allowing the model to adapt its detection thresholds based on specific requirements.

 A threshold range of 0.005–1.0 was applied to the following classes: ``green construction wall", ``dark green wall", and ``construction wall". These objects are challenging to detect due to their subtle textures and colors that often blend into the environment. Lowering the threshold allows the model to pick up more instances of these objects, ensuring they are not overlooked during detection.

A threshold range of 0.03–1.0 was used for the following classes: ``construction barricade", ``red traffic barricade", ``white traffic barricade", ``car", ``bench", ``tree", and ``building". These objects typically fall within the expected confidence range of YOLO-World's scoring system, where confidence scores often stay under 0.1 \cite{yolo_world_prompting_tips}. This threshold ensures these objects with moderate detection confidence are captured effectively without compromising precision.

A threshold range of 0.12–1.0 was used for the following classes:  ``red traffic barrier", ``orange traffic barrier" and ``traffic cone". These objects are more likely to generate false positives due to visual similarities with unrelated objects like fire hydrants, trash bins, or red tiles on sidewalks. By setting a relatively high confidence threshold, the model filters out low-confidence predictions, reducing false positives and improving overall detection accuracy.

By strategically defining and integrating redundant classes based on common misclassifications and customized thresholds for different classes, models can enhance their decision boundaries and reduce confusion among similar objects, ultimately leading to better performance in real-world applications.

\subsection{Construction Sign Detection}

Text is an integral part of urban environments, commonly found on street signs, advertisements, murals, and shop names. For pBLV, interpreting these texts presents unique challenges. Construction sites, in particular, often include critical signage that conveys important safety information, such as warnings, directions, or restricted access notices. To assist pBLV in interpreting construction signs, we employ OCR technology.

We selected PaddleOCR \cite{paddleocr} for this project, as it is one of the most reliable OCR tools available. Numerous studies have demonstrated its effectiveness in various applications, including license plate recognition \cite{reddy2024license}, health code recognition \cite{li2023research}, and the detection of urban scene texts \cite{patil2024optimized}. PaddleOCR's robust performance makes it an ideal candidate for identifying construction-related text in dynamic and challenging urban environments.

Despite its strong text detection, not all text detected in construction environments is relevant. For instance, advertisements, graffiti, or unrelated street signs can appear in the OCR output, creating noise that complicates the identification of critical construction-related information. To address this, we implemented a filtering mechanism, proposed previously  \cite{feng2023commute}, to refine the PaddleOCR output to focus on relevant signs.

To effectively filter relevant content, we developed a comprehensive dictionary of possible construction signs. This dictionary serves as a reference to verify the relevance of the text detected by PaddleOCR. Six student volunteers contributed to this effort by surveying approximately 50 construction environments, either in person or using Google Maps StreetView. Through these surveys, 171 construction signs were observed, and 64 unique strings relevant to construction were extracted. Examples of these strings include ``Authorized Personnel Only”, ``Caution: Construction Zone”, and ``Road Work Ahead”.

The string similarity was calculated to compare the OCR detected text with the entries in the construction sign dictionary. This method leverages the Sorensen-Dice coefficient \cite{kondrak2005n}, a metric that effectively measures string similarity by comparing shared character sequences. For every text output by PaddleOCR, a similarity score was calculated against all entries in the dictionary. A threshold of 0.8 was set, meaning that any text with a similarity score exceeding this threshold was considered a match to a construction-related sign.

This scoring mechanism ensures that minor variations in wording, caused by OCR inaccuracies, abbreviations, or changes in phrasing, do not prevent the accurate detection of relevant construction text. For example, even if a detected string reads ``roadwrk Ahead”, the similarity score calculation would correctly identify it as matching the intended sign, ``Road Work Ahead”. 

.

\subsection{Experimental Setup}

We conducted experiments to evaluate the system's effectiveness in outdoor environments. Four volunteers wore a smart wearable device designed for pBLV, the Visually Impaired Smart Service System for Spatial Intelligence and Navigation (VIS\textsuperscript{4}ION), to collect video footage around construction sites from various angles. VIS\textsuperscript{4}ION is a personal mobility solution offering a customizable, human-in-the-loop, sensing-to-feedback platform that provides real-time functional assistance \cite{beheshti2023smart, ng2022real,hao2022detect,9730919}. The system is a wearable backpack equipped with multiple sensors, including two cameras mounted on the shoulder straps. The cameras are Arducam 1080P Low Light Ultra Wide Angle USB Cameras \cite{Arducam2024}, recording video at 30 fps with a resolution of 1920 x 1080. The cameras have a diagonal field of view (FOV) of D = 160°.

The experimental setup consisted of two parts: a static test and a dynamic test, both conducted in NYC to evaluate the system's detection capabilities for construction site objects.

\subsection{Static Testing}

\begin{figure}[ht]
\centering
\includegraphics[width=0.6\textwidth]{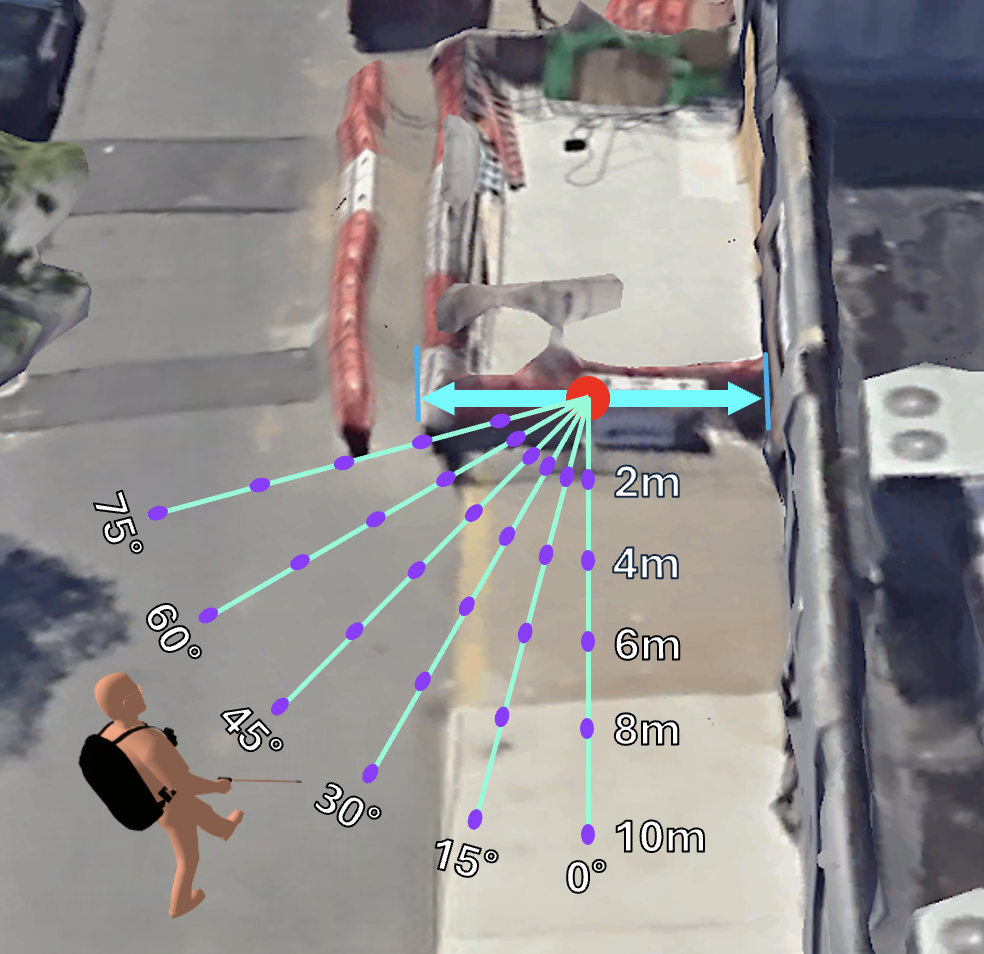}
\caption{
A graphical illustration of the data collection procedure at a construction site. The blue arrowed line represents the width of one end of the construction site, with the red point at its center serving as the reference point. Five green lines indicate measurement lines from different angles, with their angles labeled in white text. Purple dots mark the locations where the volunteer pauses for 5 seconds. An orange human figure wearing the VIS\textsuperscript{4}ION system is positioned near the 45° and 10-meter mark. This figure is for conceptual illustration purposes; the human figure's scale does not accurately reflect the actual scale of the environment.
}
\label{fig:static_test_plan}
\end{figure}

In the static testing, four volunteers visited seven distinct construction sites across NYC streets, each characterized by specific construction elements. Sites 1 through 7 prominently featured sidewalk sheds, traffic barrels, barricades, traffic cones, construction site dividers/temporary walls, construction signs, and a combination of multiple construction elements, respectively. These sites represent the most commonly observed types of construction sites in NYC. At each location, volunteers collected videos from various angles and distances. The procedure for video collection is illustrated in Figure \ref{fig:static_test_plan}. At the construction site, the width of one end of the site was measured, as shown by the blue arrowed line in Figure \ref{fig:static_test_plan}. The center of the blue arrowed line was designated as the reference point for angle and distance measurements.

A virtual line, perpendicular to the blue arrowed line and passing through the reference point, was defined as the 0° line, shown as the green line labeled 0° in Figure \ref{fig:static_test_plan}. The 15° line was defined as a line passing through the reference point, forming a 15° angle with the 0° line. Similarly, the 30°, 45°, 60°, and 75° lines were defined in the same manner, as shown in the green lines with angle labels in Figure \ref{fig:static_test_plan}. Along each line, measurement points were marked at distances of 2, 4, 6, 8, and 10m from the reference point. Some of these points are depicted as purple dots in Figure \ref{fig:static_test_plan}. Our volunteers recorded videos by walking along each virtual line and pausing at each purple dot for at least 5 seconds to ensure clear video capture. After recording, frames at each measurement point were extracted for evaluating purposes.

\subsection{Dynamic Testing}

Dynamic testing is essential for evaluating the reliability and performance of systems in real-world urban walking scenarios. Unlike static testing that merely assesses the detection range using clear video frames, dynamic testing addresses additional ecologically valid complexities, such as motion blur, object occlusion, and varying lighting conditions, which influence the system’s robustness. Dynamic testing simulates real-world scenarios that systems will face daily, making it crucial for evaluating their effectiveness and reliability.

In this testing, one volunteer wore the VIS\textsuperscript{4}ION system and walked naturally along NYC streets. A volunteer completed a trip in the Downtown Brooklyn area, covering approximately 0.5 miles in about 10 minutes. The video data collection was conducted at the volunteer's own comfortable walking pace. The recordings took place on a typical sunny summer afternoon in NYC, with average summer temperature, humidity, and sunlight conditions. No abnormal wind, noise, or shadow conditions were reported during the recording. This route encountered 8 different construction sites. Approximately one-fourth of this route required the volunteer to navigate areas impacted by construction sites. These included walking under sidewalk sheds, traversing narrower paths created by barricades, and rerouting due to road closures, etc. The non-construction sections of this route are set in a typical Downtown Brooklyn business area. Key features include encountering other pedestrians every few steps, illegally parked cars partially blocking the path, and trash piles scattered on the sidewalks. According to NYC Open Data \cite{sidewalk_widths_nyc}, the sidewalk width along this route ranges from approximately 2.5 meters to 4.8 meters.

\subsection{Evaluation Metrics for the Static Testing}

In static testing, 30 unique frames were extracted at a construction site, each frame was extracted from a measurement point, representing a specific view based on angle and distance, such as 0° at 2m, 0° at 10m, 75° at 6m, and so on. For each measurement, at least 5 seconds of video were recorded, generating multiple frames. The extracted frame was selected as the first frame where the camera stabilized, and the image became clear. These frames were used to evaluate the system's performance in static conditions using three key metrics: accuracy, angular offset, and distance. 

\subsubsection*{Accuracy}

Accuracy measures the overall success rate of identifying construction-related objects or signs within a construction site. Since testing frames are extracted from measurement points, a construction site must exist within the frame. A successful detection is when the proposed model also classifies these frames as containing a construction site.

The accuracy across the tested range is calculated using the following formula:

\[
\text{Accuracy} = \frac{\text{Number of Successful Detected frames}}{\text{Total Number of Frames}} \times 100\%
\]

This metric serves as a crucial indicator of the model's overall accuracy and reliability in object detection.

\subsubsection*{Angularoffset}

Angular offset is a critical metric used to evaluate a system’s ability to detect objects at varying angles. It reflects the effectiveness of the system in identifying construction-related objects or signs when observed from different angles.

\[
\text{Angular offset} = \frac{\text{Number of Frames at a Given Angle with Successful Detections}}{\text{Total Number of Frames at that Angle}} \times 100\%
\]

This metric is determined by calculating the ratio of successful detections at a specific angle to the total number of frames tested at that angle. The definition of successful detection is the same as above, that the predominant construction object(s) in that frame is detected. At each construction site, angular-Offset was measured across five detection angles: 0°, 15°, 30°, 45°, 60°, 75°.

\subsubsection*{Distance}

This metric is determined by calculating the ratio of successful detections at a specific distance to the total number of frames tested at that distance. The definition of successful detection remains the same, requiring that the predominant construction object(s) in a frame be detected. At each construction site, distance was measured across five distance: 2m, 4m, 6m, 8m and 10m.

This metric is crucial for understanding how detection performance correlates with distance, helping to evaluate how varying ranges impact the system’s ability to detect objects.

\subsection{Evaluation Metrics for the Dynamic Test}

In the dynamic test scenario, we relied on several performance metrics to gauge the model’s ability to classify each video frame accurately. These metrics are defined based on how well the system distinguishes between ``construction site" and ``not construction site" frames.

Accuracy is a fundamental measure of the proportion of frames correctly classified. It reflects the overall reliability of the system in identifying construction-related frames while avoiding misclassifications of non-construction frames.

Error rate is determined by the proportion of frames that are misclassified, providing an inverse perspective on the model’s performance. A higher error rate signals a greater number of incorrect classifications.

Precision focuses on frames predicted as ``construction site", indicating how many of these predictions are actually correct. It provides insight into the model’s ability to minimize false alarms.

Recall measures the fraction of true construction-site frames that the model successfully identifies. This metric highlights the model’s capability to detect all relevant instances of construction activity.

F1 Score represents the harmonic mean of Precision and Recall, offering a single measure that balances these two aspects of performance. It is particularly useful when seeking to optimize both the identification and correctness of “construction site” predictions.

Specificity illustrates how effectively the model identifies non-construction frames. It reflects the model’s accuracy in filtering out frames that do not contain construction activity, ensuring fewer false positives.

\section{Results}

\subsection{Static Testing}

The overall accuracy across all measurement points and all seven construction sites includes 210 measurement points in total and results in 186 correct detections (88.56\%). The breakdown of accuracy at each measurement point is shown in Table \ref{tab:distance_angle}.

\begin{table}[h!]
\centering
\caption{Detection Success Rates at Different Distances and Angles}
\begin{tabular}{cccccccc}
\toprule
\textbf{Distance/Angle} & \textbf{0°} & \textbf{15°} & \textbf{30°} & \textbf{45°} & \textbf{60°} & \textbf{75°} & \textbf{Mean} \\
\midrule
2 m  & 100\% & 100\% & 100\% & 100\% & 100\% & 100\% & 100\%  \\
4 m  & 100\% & 100\% & 100\% & 100\% & 100\% & 100\% & 100\%  \\
6 m  & 86\%  & 100\% & 100\% & 100\% & 100\% & 86\%  & 95.3\% \\
8 m  & 86\%  & 86\%  & 100\% & 86\%  & 86\%  & 57\%  & 83.5\% \\
10 m & 86\%  & 86\%  & 57\%  & 57\%  & 57\%  & 43\%  & 64.3\% \\
\midrule
\textbf{Mean} & 91.6\% & 94.4\% & 91.4\% & 88.6\% & 88.6\% & 77.2\% & 88.6\%  \\
\bottomrule
\end{tabular}
\label{tab:distance_angle}
\end{table}

The angular offset was evaluated at incremental angles from 0° to 75° across the seven sites. The mean success rates (±SD) were as follows: 0°, 91.4\% (±7.0\%); 15°, 94.3\% (±7.0\%); 30°, 91.4\% (±17.1\%); 45°, 88.6\% (±16.7\%); 60°, 88.6\% (±16.7\%); and 75°, 77.1\% (±23.2\%). These results demonstrate that the detection accuracy is highest within the range of 0° to 30°, consistently above 90\%. However, as the angle increases beyond 30°, the accuracy shows a gradual decline, with a notable drop to 77.1\% at 75°. The higher accuracy at smaller angles suggests that the system performs best when objects are closer to the frontal view.

The distance was evaluated at incremental distances from 2 m to 10 m, yielding mean success rates (±SD) of 100\% (±0\%) at 2 m, 100\% (±0\%) at 4 m, 95.2\% (±6.7\%) at 6 m,  83.3\% (±12.8\%). at 8 m, and  64.3\% (±16.0\%). at 10 m. Overall, these results demonstrate a consistent decrease in detection coverage with increasing distance, suggesting that the model’s performance is robust at closer ranges but becomes progressively challenged as the target objects recede from the camera.

\begin{figure}[ht]
\centering
\includegraphics[width=0.8\textwidth]{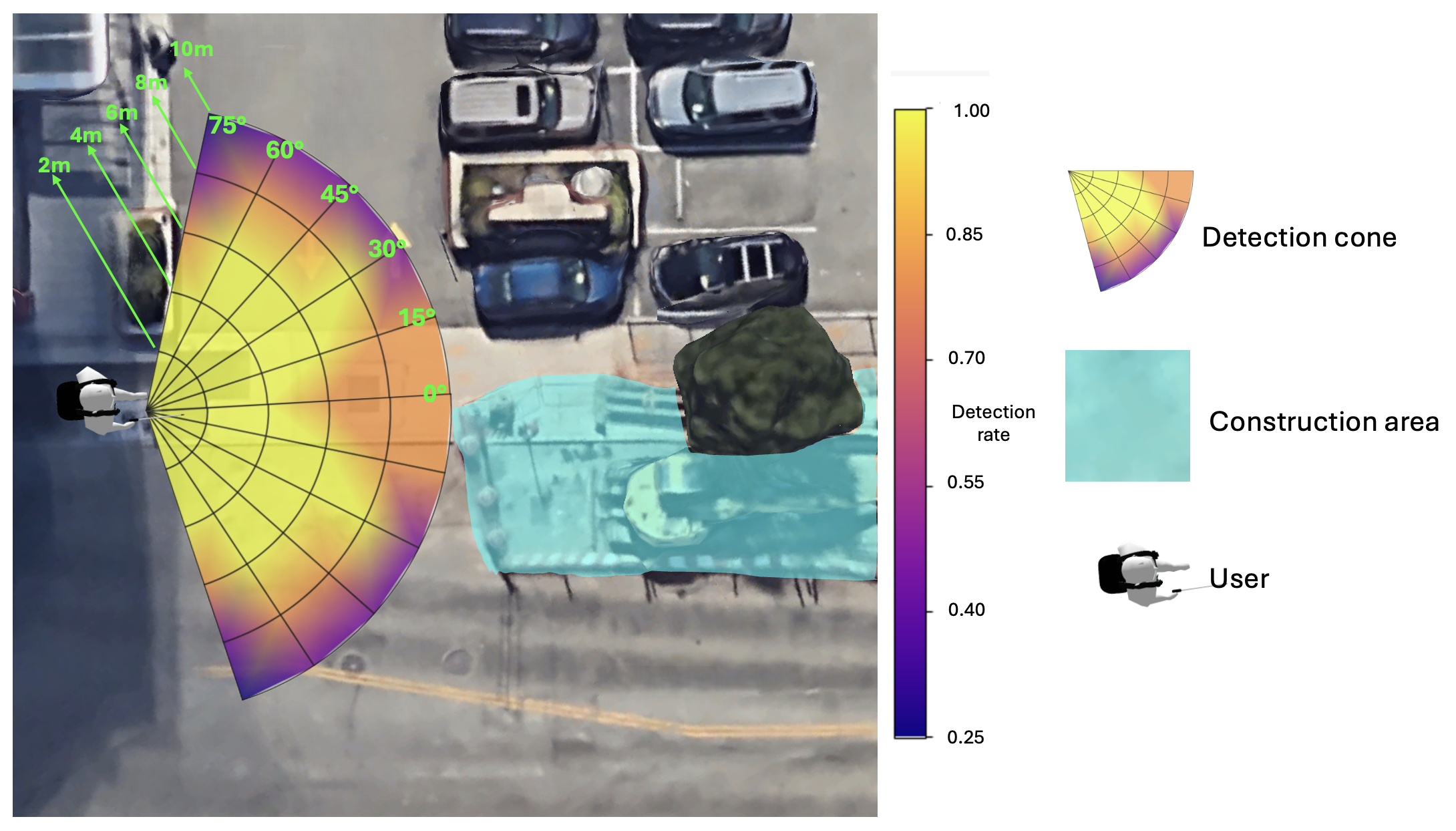}
\caption{The visualization of the detection range. The colored fan-shapes depict detection success rates at different angles and distances: yellowish areas indicate higher success rates, while purplish areas show lower rates. The green colored area represents the construction area. }
\label{fig:range_result}
\end{figure}

This detection range is illustrated in Figure \ref{fig:range_result}, where a color map visualizes the detection success rate at various locations. A yellowish color indicates a high accuracy, while a purplish color represents areas with a lower accuracy. From Figure \ref{fig:range_result}, it is evident that accuracy is higher when the objects are close to the user or directly in front of them.

Detection performance varies based on the predominant features of the construction site. For construction sites characterized by wall dividers, traffic barrels, or traffic cones, the system successfully detected objects across all measurement points. The accuracy at these sites was 100\% (±0\%). The angular offset reached 100\% across the entire tested range, from 0° to the maximum testing distance of 10 m, where the accuracy remained 100\%.

For construction sites with predominant features of sidewalk sheds, detection failures occurred only at 10 m and 8 m for the 75° angle. This indicates that 75° is a particularly challenging angle for the model to detect objects correctly at greater distances.

For construction sites dominated by traffic barricades, unsuccessful detections were observed at 10 m for the 30°, 45°, 60°, and 75° angles, at 8 m for the 45° and 75° angles, and at 6 m for the 75° angle. Overall, this type of construction site proves more challenging to detect at a distance of 10 m or at the 75° angle.

For OCR-based reading, successful text recognition occurred only within 4 meters. Within this range, detection was consistently successful across all angles.

\subsection{Dynamic Testing}

We collected a test video containing approximately 14,886 frames, of which 3,793 were labeled as construction-site frames and 11,093 were labeled as non-construction-site frames.

Each frame was processed through the decision-making framework to determine whether it was correctly classified as a construction site. The results showed 86.30\% accuracy, 71.64\% precision, 76.64\% recall, and 74.0\% F1 score. Results are shown in Table \ref{tab:confusion_matrix}.

\begin{table}[h!]
\centering
\begin{tabular}{lcc}
\toprule
\textbf{Actual / Tested} & \textbf{Positive} & \textbf{Negative} \\
\midrule
\textbf{Positive}        & 2907              & 886               \\
\textbf{Negative}        & 1151               & 9921              \\
\bottomrule
\end{tabular}
\caption{Confusion matrix for the dynamic testing}
\label{tab:confusion_matrix}
\end{table}

\subsubsection{K-Frame Majority Voting}

One way to improve the output is K-Frame Majority Voting, a post-processing technique used in temporal data analysis, particularly in video sequences, to enhance accuracy and consistency of predictions across consecutive frames \cite{lam1997application}. The idea is to use predictions from K consecutive frames and decide the final output for a specific frame based on the majority class or decision within those frames.  We tested various K values, and the results showed that accuracy and F1 Score steadily increased as K grew from 10 to 50 frames. For example, at 10 frames, the accuracy was 0.89 and the F1 Score was 0.80, while at 50 frames, the accuracy improved to 0.91 with an F1 Score of 0.82. Beyond 50 frames, the improvement in metrics became marginal, with 60 and 100 frames both achieving an accuracy of 0.91 and an F1 Score of 0.83.

Based on these results, 50-frame majority voting provides the best balance between accuracy (0.91) and computational efficiency, as increasing K beyond this point yields diminishing returns. Therefore, we conclude that 50-frame majority voting is the most effective and practical method for this application.

\section{Discussion}

The results demonstrate that this proposed method effectively detects construction site objects with high accuracy. Across all tested angles and distances, the model achieved an overall accuracy of 88.56\%, correctly identifying 186 out of 210 measurement points. Detection success rates remained high within a 0° to 30° viewing angle, exceeding 90\%, but gradually declined beyond 30°, reaching 77.1\% at 75°. Distance also impacted detection performance, with success rates at 100\% for objects within 4 meters, 95.2\% at 6 meters, 83.3\% at 8 meters, and 64.3\% at 10 meters. Object type influenced detection outcomes—construction dividers, traffic cones, and traffic barrels exhibited the highest detection success across all tested conditions, while traffic barricades and signs showed reduced detection reliability, particularly at oblique angles or greater distances. In dynamic testing, the model achieved an accuracy of 86.30\%, with an F1 score of 74.0\%. Performance improved when applying K-frame majority voting, reaching a peak accuracy of 91\% at 50-frame aggregation.

Object detection performance also varied with the user's viewing direction. At a 0° angle, all construction sites—except those with text-based signs—were detected up to 10 meters. Objects with distinctive colors and textures, such as green construction walls, were reliably detected at extreme distances and angles. The strong representation of traffic cones and barrels in the YOLO-World pre-trained dataset contributed to their higher confidence scores and improved detection accuracy \cite{roboflowYoloWorldPromptingTips}. In contrast, objects like traffic barricades and construction signs posed greater detection challenges, particularly when viewed at oblique angles.

Additionally, detection accuracy was influenced by object positioning. When an object was angled away from the user, detection success required closer proximity. This is particularly relevant in real-world applications, as users may need to adjust their walking path or move closer to certain objects before they can be reliably identified. Notably, the ability to detect construction sites at 10 meters suggests that pBLV users could make rerouting decisions before reaching hazardous areas, improving their overall navigation safety.

Moreover, one key challenge observed was false positives in open vocabulary object detection due to overgeneralization. For instance, the model frequently misclassified objects with similar visual features, such as mistaking streetlight poles or silver window frames for scaffolding poles. To mitigate this, we developed a specialized YOLO model focused on scaffolding pole detection, which significantly reduced false positives and improved precision.

In the next sections, we further analyze the causes of detection inaccuracies, including specific challenges related to object orientation, occlusions, and model confidence thresholds.

\subsection{Type of Errors}
\subsubsection{False Positives}
\paragraph{A. Constructions in the Background}

\begin{figure}[ht]
\centering
\includegraphics[width=0.5\textwidth]{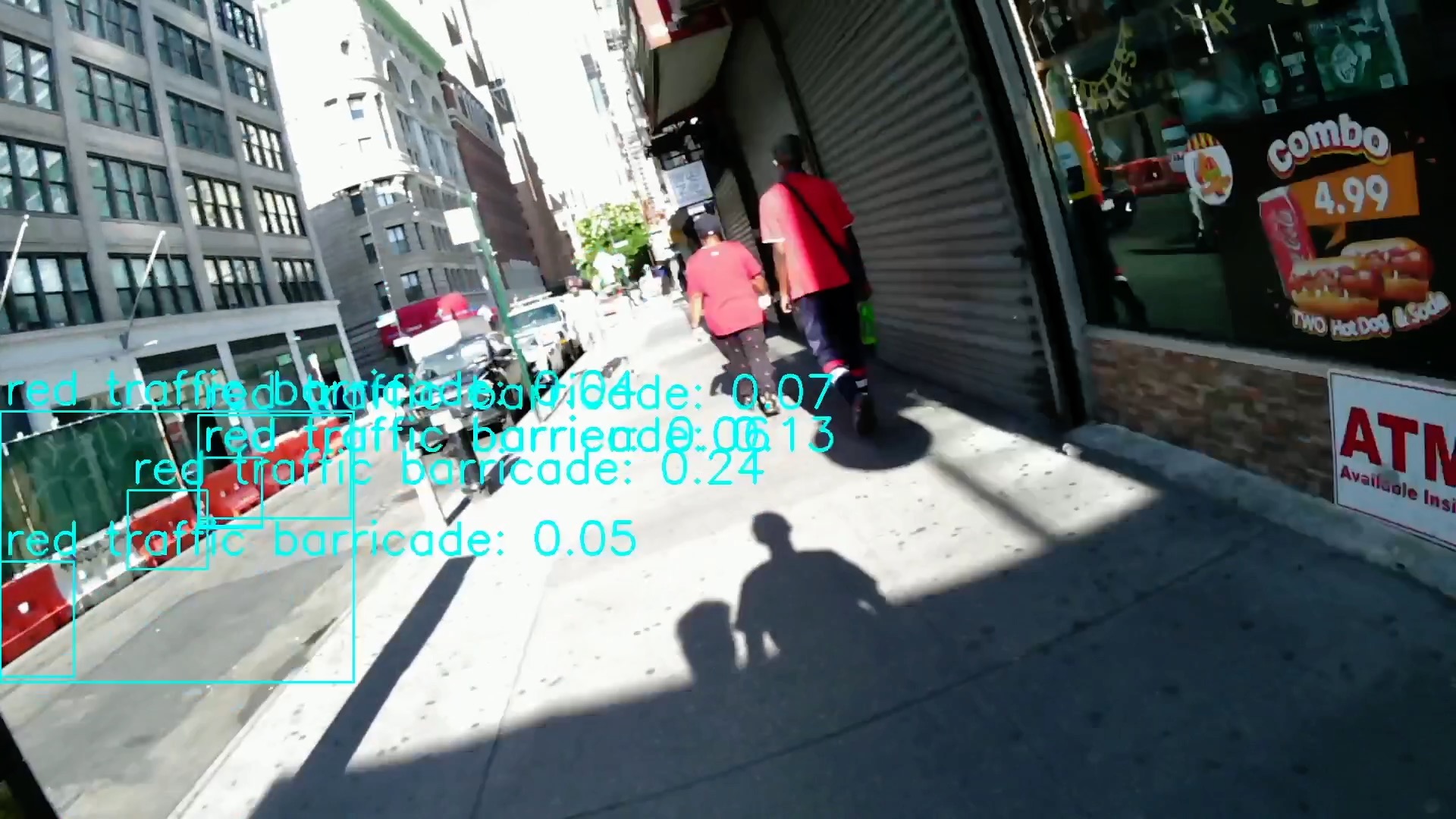}
\caption{An example of false positives arises from the construction sites located on the opposite side of the walkway.}
\label{fig:other_side}
\end{figure}

In dynamic testing, false positives were primarily triggered by construction sites on the opposite side of the walkway, as shown in Figure \ref{fig:other_side}. Therein, the construction area occupies both the sidewalk and the adjacent vehicle lane, resulting in some construction-related objects being positioned near the opposite sidewalk from where the recorder is walking. These construction-related objects fall within the detection range of the proposed model, which triggered a positive detection result.  

However, such false positives are triggered only under extreme circumstances. In our testing, these errors were most likely to occur when the construction site was on a narrow road, partially occupying the vehicle lane, and there was no traffic or parked vehicles obstructing the view. Notably, in this testing, all such false positives occurred on one-way roads. Construction on the opposite side of two-lane roads never triggered this type of false positive. Although construction on the opposite sidewalk is common, the specific conditions required to trigger this false positive are rare.

\paragraph{B. Irrelevant Construction Objects}

\begin{figure}[ht]
\centering
\includegraphics[width=0.4\textwidth]{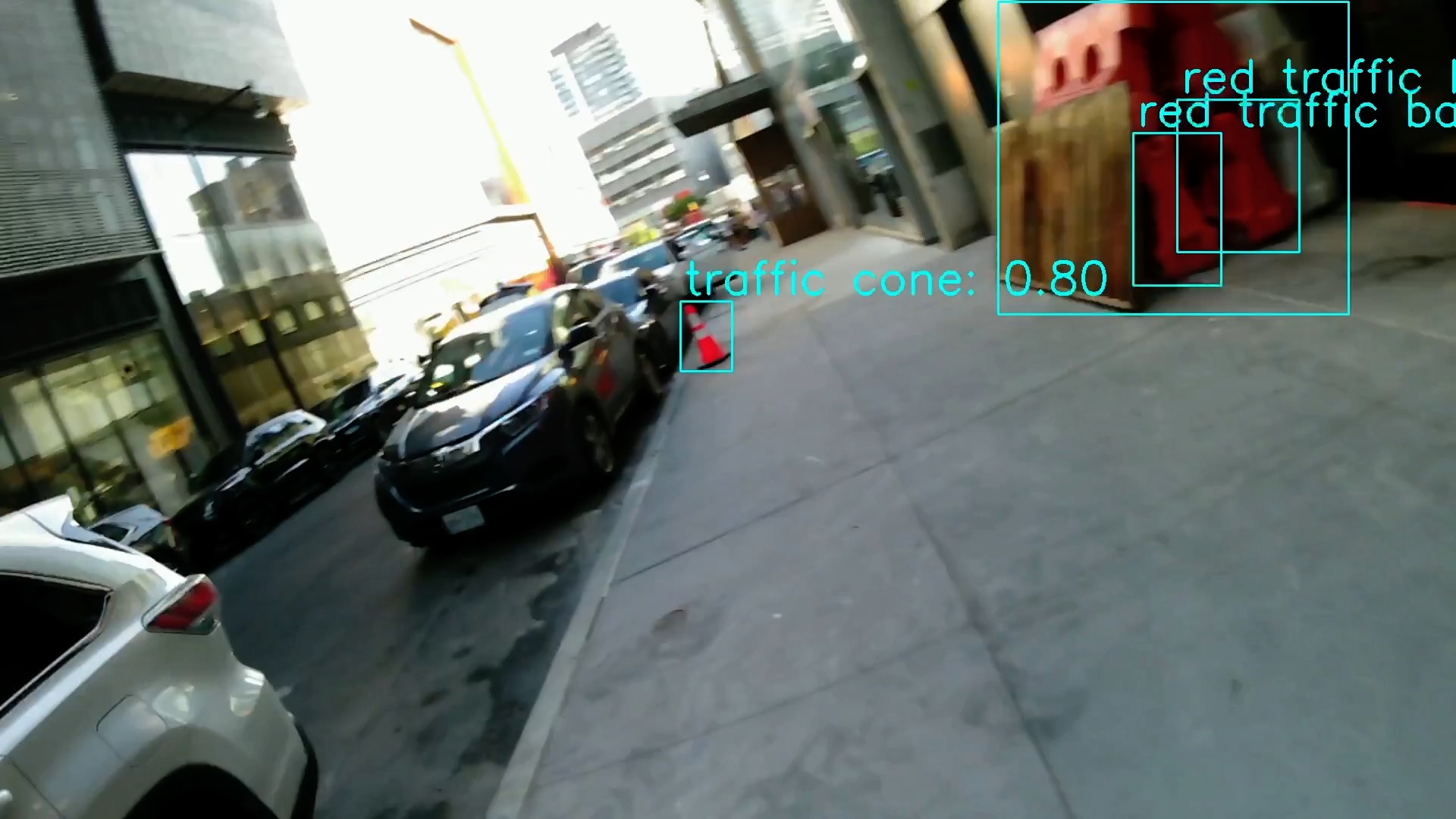}
\caption{A frame showing construction-related objects successfully detected; however, these objects are not associated with active construction sites.}
\label{fig:no_construction}
\end{figure}

In real-world environments, construction-related objects may be present without actively serving construction purposes. As shown in Figure \ref{fig:no_construction}, several traffic barricades are placed along the side of the road. These barricades appear to be out of use, likely stacked together after construction work was completed. Additionally, a traffic cone on the left side of the sidewalk seems to have been unintentionally left there, with no apparent hazards in the area.

The presence of these stacked barricades results in a high number of detected objects, leading the system to erroneously conclude the existence of a construction site. While these objects are indeed construction-related and their detection is technically accurate, their accumulation in a small area does not affect the walkability of the road. Consequently, these instances are classified as false positives.

Future improvements to address this issue could leverage methods such as the Relation Network \cite{hu2018relation}, a deep learning architecture that explicitly models object-to-object relationships within an image. By analyzing the spatial relationships between construction objects and their connection to the ground, the system could better differentiate between actively used objects and those that are merely stored or misplaced.

\subsubsection{False Negatives}

\paragraph{ A. Motion Blur}

\begin{figure}[ht]
\centering
\includegraphics[width=0.4\textwidth]{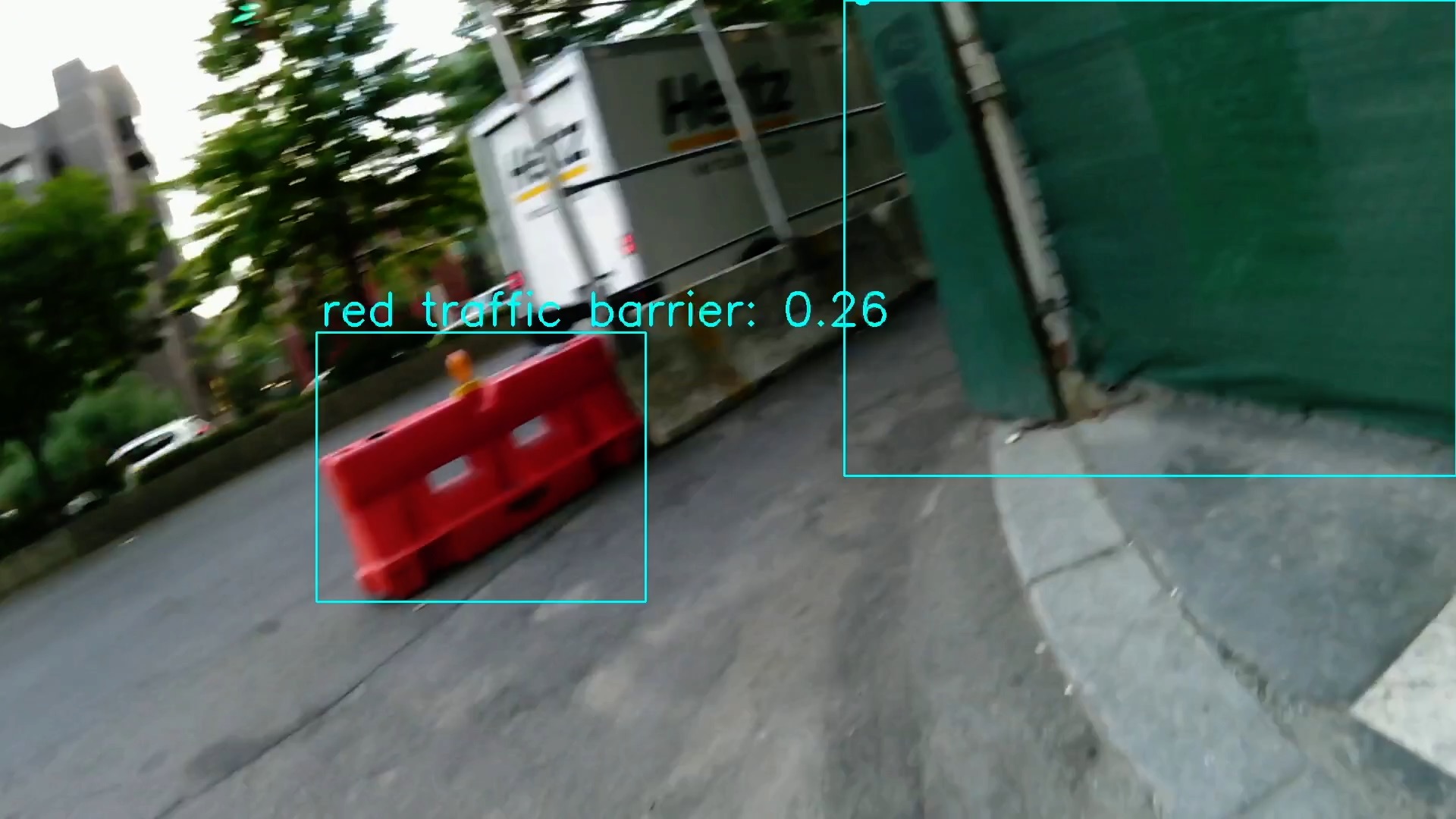}
\caption{Detection results for traffic barricades at varying distances and angles. Green dots indicate successful detections.}
\label{fig:motion_blur}
\end{figure}

Motion blur can sometimes cause problems with detection. Even though a frame rate of 30 frames per second works well for most walking scenarios, it struggles when the user makes sharp turns. During a turn, the angular velocity becomes high, and this creates motion blur in the video. As shown in Figure \ref{fig:motion_blur}, motion blur during a turn can make it difficult to detect scaffolding poles. For example, in the figure, the closest scaffolding pole is only about 4 meters away, yet it is not detected. This is in contrast to static testing, where the pole at same distance is successfully detected. Once the user finishes the turn and resumes walking in a straight line, the detection returns to normal. In Figure \ref{fig:motion_blur}, Only a traffic barricade and a green construction divider were detected, which is insufficient to meet the threshold required to be considered a positive detection.

To address this challenge, potential solutions could include integrating an Inertial Measurement Unit (IMU) to detect turns and temporarily ignore frames captured during high angular velocity. Additionally, implementing a quality metric, such as Peak Signal-to-Noise Ratio (PSNR) \cite{poobathy2014edge}, a metric used to measure the quality of an image or video by comparing it to a reference version, could ensure that only high-quality frames are processed, reducing the likelihood of false negatives caused by motion blur. Another possible solution is a hardware upgrade to use a high-frame-rate camera. Capturing more frames per second improves motion accuracy, resulting in smoother visuals and less blurring of fast-moving objects. These combined measures would significantly enhance the robustness and reliability of the detection system under dynamic conditions.

\paragraph{B. Unexpected Change in Construction Objects Appearances}

\begin{figure}[ht]
\centering
\includegraphics[width=0.4\textwidth]{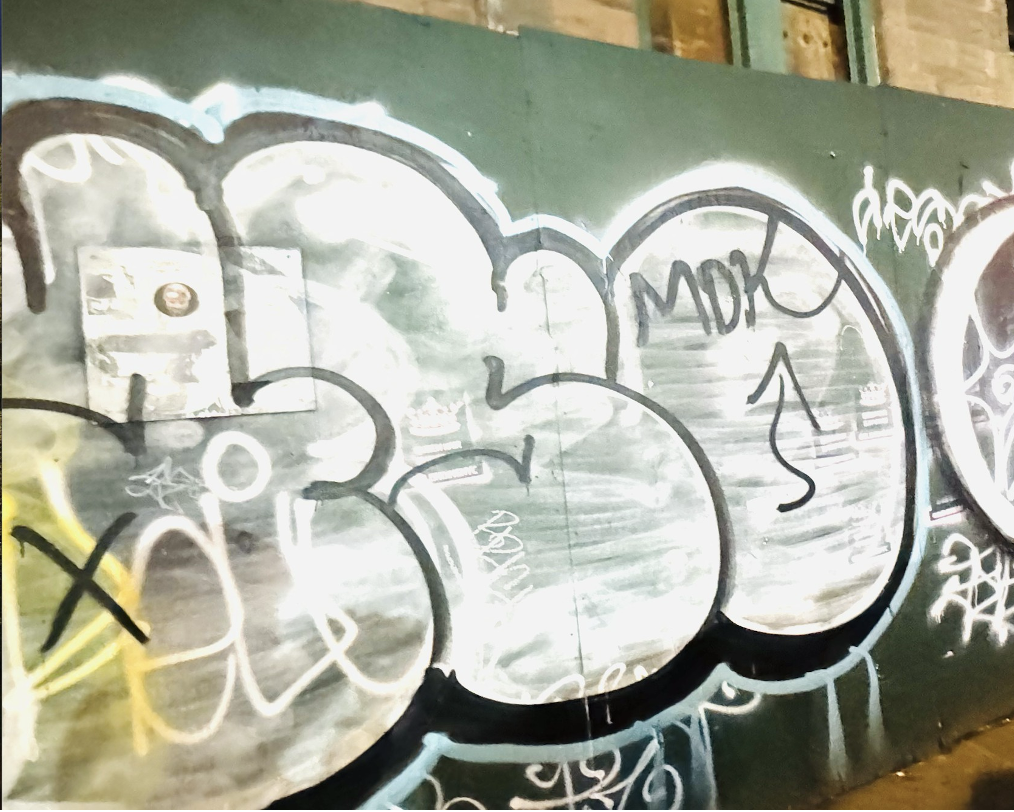}
\caption{An example of false negatives arises from a green wall covered by graffiti}
\label{fig:construction_wall_with_painting}
\end{figure}

Another cause of false negatives is the unexpected alteration in the appearance of construction objects. For instance, as shown in Figure \ref{fig:construction_wall_with_painting}, a construction wall divider became completely covered in colorful graffiti. While graffiti is technically illegal \cite{Tablet2024, NYC2017}, it is widespread in urban environments, transforming the appearance of these walls. This transformation replaces the original uniformity of the walls with irregular and complex patterns, obscuring the key visual features that detection systems rely on for accurate recognition. Consequently, these visual alterations often prevent the system from correctly identifying construction sites, leading to false negatives.

To mitigate the impact of such visual changes, several strategies can be implemented. One approach is to train detection models on a more diverse dataset that includes examples of construction objects with graffiti and other unexpected alterations. By exposing the model to a wider range of appearances, its ability to generalize and accurately recognize construction walls under varied conditions can be significantly enhanced.

Additionally, integrating texture and shape analysis into the detection algorithm can help address this challenge. Rather than relying solely on color or patterns, the algorithm can focus on structural features such as the rectangular shape and dimensions of construction walls. This would enable the system to identify construction objects based on their inherent geometry, even when their surface appearance is altered.

\paragraph{C. OCR Error}

\begin{figure}[ht]
\centering
\includegraphics[width=0.4\textwidth]{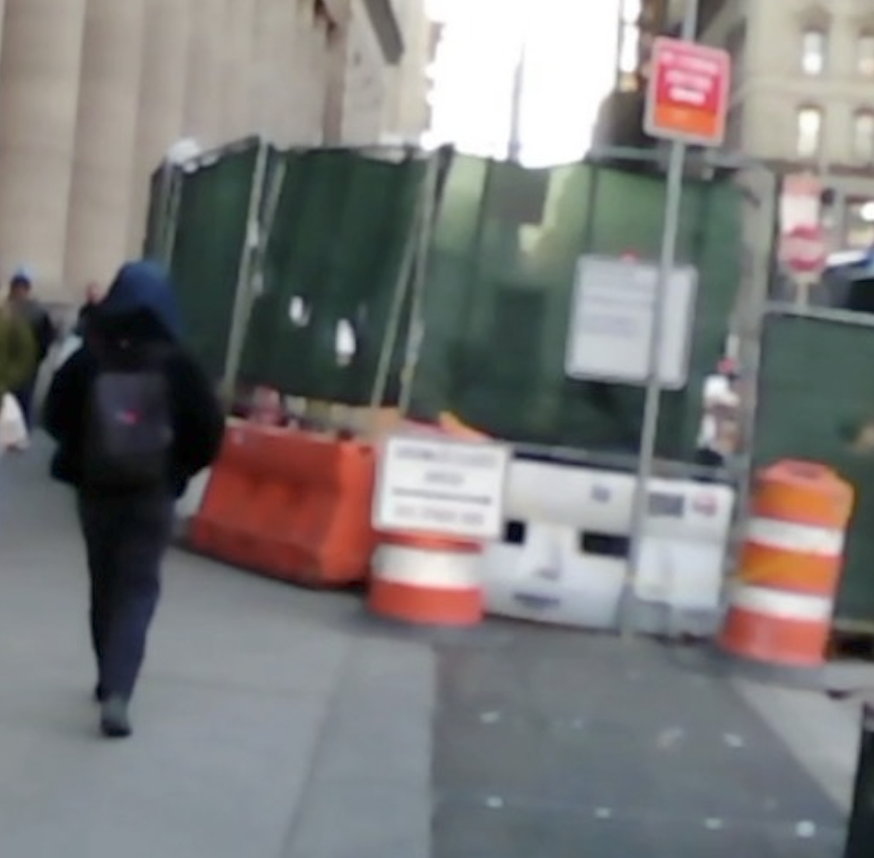}
\caption{A cropped photo of a construction sign taken from 10 meters away. }
\label{fig:OCR_error}
\end{figure}

Construction sign detection can also result in false negatives. This error is often caused by text that is unrecognizable to the OCR model. As shown in Figure \ref{fig:OCR_error}, which is a cropped frame taken from 0° and 10 meters away, the text on the sign is not recognizable, even to the naked eye. In this frame, a construction sign with black font on a white background is attached to a traffic barrel. The sign contains four rows of text: the first row reads ``Sidewalk Closed", the second row says ``Ahead", the third row features an arrow pointing to the left side of the image, and the fourth row reads ``Use Other Side".  However, due to the small font size on the sign and the blurriness of the image, the text is not recognizable. As a result, the OCR model is unable to read the construction sign from any angle when the distance exceeds 8 meters. At 6 meters, only half of the angles result in successful detection, while within 4 meters, the OCR model can successfully recognize the sign from all angles.

To address this issue, a camera upgrade with auto-focusing capabilities could be considered. Currently, the camera used in this project has a manually adjustable focal point, which is the primary cause of the blurriness in this frame. It is not feasible to manually adjust the focus for every object. An auto-focusing feature would likely resolve this problem. Additionally, implementing a construction sign detector could further mitigate this issue. By detecting the location of the construction sign, zooming in on that area, and then applying the OCR model, the system would have a higher likelihood of successfully recognizing the text.

\subsection{Limitations}
\subsubsection{Limitations in Experiments}
All testing images were collected on sunny or cloudy days under typical daylight conditions. This setup limits the range of environmental factors that might impact the detection results, such as varying lighting conditions.

Moreover, the performance under occlusions was not evaluated. It is common for target objects to be partially occluded by other pedestrians, cars, etc. The extent to which occlusion affects recognition failure is unknown.

Furthermore, the generalizability of the approach requires more testing. While this experiment demonstrated potential in using a text-prompt approach for detecting various construction sites, the generalizability of this method warrants further investigation. For example, how this system performs in another city with construction-related objects of different colors remains unexplored.

\subsubsection{Limitation in Open Vocabulary Object Detection }

While open vocabulary object detection opens a new door for construction site detection, and this study showed this as a promising solution as part of a larger pipeline of AI approaches, it still faces some limitations.

One limitation, as mentioned earlier, is overgeneralization. Without any techniques to mitigate it, the detection accuracy is much worse than that of a YOLO model trained on a well-labeled dataset.

Another limitation is understanding the viewpoint dependency. For the class of barricades, it can be detected at 10 meters away at angle of 0° and 15°, but the detection range reduced to 4 meters at 60° and 75°. But for traffic barrel or traffic cone, such issue is not observed. The significant shift in visual characteristics between the side and front views of barricades affects their detection capabilities. For example, the front view appears as a flat rectangular panel, while the side view takes on an “A” shape, which complicates detection at various angles. The traffic barrel or traffic cone look identical no matter from side view or front view, which causes no change in detection distance at different angles.

One strategy to mitigate false positives was the introduction of redundant null classes. This approach involves adding classes that absorb misclassifications, thereby decreasing the likelihood of incorrectly identifying irrelevant objects. For instance, we identified common misclassification pairs, such as green construction walls being mistaken for grassland. However, identifying these null classes is a time-intensive process and inevitably leaves some classes unaddressed, leading to residual inaccuracies.

These challenges underscore the need for more refined techniques to enhance the model’s ability to distinguish visually similar objects. While open vocabulary object detection shows immense potential for revolutionizing object detection across various domains, including construction, further development is required to ensure its reliability in real-world applications.

\subsection{Benefits of Majority Voting over Time}

We found that 50-frame majority voting improves accuracy and consistency by leveraging temporal redundancy in video sequences to filter out noise and occasional misclassifications. Random errors, such as those caused by motion blur, occlusions, or poor lighting, are often isolated to specific frames. By considering predictions from consecutive frames, the method smooths out these errors. Since objects typically remain visible across multiple frames, temporal redundancy ensures that even if one frame is misclassified, surrounding frames correct it.

Majority voting also aggregates predictions, creating a more stable and reliable output. Instead of abrupt classification changes due to isolated errors, the final result becomes smoother and more consistent over time.

However, while increasing K improves stability, it also impacts inference time. Higher values of K require processing and storing predictions from a larger number of frames before making a final decision, which could introduce latency in real-time applications. In our tests, we found that K=50 achieves an optimal balance between accuracy and efficiency. Beyond this threshold, improvements in accuracy become marginal, while the additional computational overhead may slow down responsiveness from the user’s perspective. Thus, selecting an appropriate K involves balancing detection reliability with real-time performance constraints.

\subsection{Future Directions}

Future improvements to the system focus on both technical enhancements and refined detection methods. Currently, the model may still misclassify certain non-harmful objects as construction sites. Addressing this issue will involve improving the algorithm’s ability to identify features such as road lanes, which can help eliminate false positives that appear across the road. Additionally, incorporating other AI models can expand the system’s functionality and improve contextual awareness. For example, visual language models could generate more accurate scene descriptions, helping the system differentiate between actual construction zones and visually similar but non-hazardous objects. Semantic segmentation models could further refine object classification by analyzing road structure and distinguishing between construction barriers and regular roadside objects. Transformer-based models for scene understanding could enhance the system's ability to recognize environmental context, reducing misclassifications caused by ambiguous visual features. Depth estimation models could be used to estimate the distance between the user and construction obstacles, enabling the system to provide more precise navigational guidance. By integrating depth information, the system could help users determine whether they can safely pass through a construction zone or if they need to reroute. By fine-tuning the detection process and integrating these AI advancements, the system will be better equipped to recognize and disregard objects that do not pose a hazard while improving overall detection reliability.

.

Another key area of advancement involves integrating real-time updates. Forging partnerships with city planning departments can facilitate the sharing of up-to-date information on active construction sites, leading to more accurate warnings and alerts. While some agencies—like the Department of Buildings in New York City—already maintain records of ongoing construction work, these datasets are often compiled passively when permits are issued or expire. Moreover, the level of impact on sidewalks is not always captured. By allowing users to record and share precise locations of obstructed walkways, the system can actively update municipal data, ensuring a live map of sidewalk construction sites. This real-time exchange of information has the potential to greatly enhance public safety and accessibility.

Finally, extensive field testing is critical to validate the system’s effectiveness in real-world scenarios. All testing so far has served as technical validation. The performance of this system with real users with low vision remains unexplored. By conducting usability studies with pBLV, we can gather valuable feedback on any remaining gaps or shortcomings in the design. This process not only refines the technology but also ensures that the solution genuinely meets the needs of those who rely on it the most.

\section{Conclusion}

This study presents a novel computer vision-based system designed to detect construction sites and their associated elements, addressing a critical gap in assistive technology for pBLV. By integrating an open vocabulary object detection model, a scaffolding-specific YOLO model, and an OCR-based construction sign recognition module, the system provides a comprehensive solution for navigating dynamic urban environments with high construction density. The system demonstrates technical feasibility and robustness, achieving high detection accuracy in controlled tests and reasonable performance in real-world scenarios.

Static testing showed 100\% accuracy for large, easily recognizable objects like traffic cones, barrels, and green construction walls across varying distances and angles. However, smaller or visually complex objects, such as scaffolding poles and construction signs, posed challenges due to angle-dependent visibility, motion blur, and camera limitations. Dynamic testing achieved an 84\% accuracy rate in distinguishing construction sites from non-construction areas, though false positives from distant construction sites and irrelevant objects highlighted areas for improvement.

These findings underscore the potential of integrating advanced computer vision techniques, such as open vocabulary detection and OCR, to enhance accessibility for pBLV. While open vocabulary detection shows great promise for assistive technology and construction site monitoring, it remains prone to false detections. Complementary approaches, such as regular YOLO models or redundant null classes, are necessary to improve accuracy and reliability. Future advancements will focus on refining detection algorithms, incorporating real-time construction data in collaboration with municipal agencies, and conducting usability testing with pBLV to further enhance independence and safety in navigating urban environments.

\section{Acknowledgment}
This research was supported by the National Science Foundation under Grant Nos. ITE-2345139, ECCS-1928614, CNS-1952180,
and ITE-2236097 by the National Eye Institute and Fogarty International Center under Grant No. R21EY033689, as
well as by the U.S. Department of Defense under Grant No. VR200130. The content is solely the responsibility of the authors and does not necessarily represent the official views of the National Institutes of Health, National Science Foundation,
and Department of Defense.

This manuscript employed ChatGPT-4o (OpenAI) for grammar checking and language refinement, while ChatGPT-O1 was utilized for code debugging. The authors carefully reviewed and validated all outputs to ensure accuracy and maintain the research’s intellectual rigor.

\section{Disclosure Statement}

New York University (NYU) and John-Ross Rizzo (J.-R.R.) have financial interests in related intellectual property. NYU owns a patent licensed to Tactile Navigation Tools. NYU, J.-R.R. are equity holders and advisors of said company.

\vspace*{ 1 cm}

\bibliographystyle{ieeetr}
\bibliography{references}

\end{document}